\documentclass[pdflatex,sn-mathphys-num]{sn-jnl}

\usepackage{graphicx}%
\usepackage{multirow}%
\usepackage{amsmath,amssymb,amsfonts}%
\usepackage{amsthm}%
\usepackage{mathrsfs}%
\usepackage[title]{appendix}%
\usepackage{xcolor}%
\usepackage{textcomp}%
\usepackage{manyfoot}%
\usepackage{booktabs}%
\usepackage{algorithm}%
\usepackage{algorithmicx}%
\usepackage{algpseudocode}%
\usepackage{listings}%
\usepackage{comment}
\usepackage{stmaryrd}

\newcommand{\might}{\lozenge}
\newcommand{\must}{\square}

\theoremstyle{thmstyleone}%
\theoremstyle{thmstyletwo}%

\theoremstyle{thmstylethree}%

\begin{document}

\title[Article Title]{A Multimodal Framework for Aligning Human Linguistic Descriptions with Visual Perceptual Data}


\author*[1]{\fnm{Jospeh} \sur{Bingham}}\email{jbingham@campus.technion.ac.il}


\affil*[1]{\orgdiv{Faculty of Biology}, \orgname{Technion University}, \orgaddress{\city{Haifa}, \country{Israel}}}


\abstract{Establishing stable mappings between natural language expressions and visual percepts is a foundational problem for both cognitive science and artificial intelligence. Humans routinely ground linguistic reference in noisy, ambiguous perceptual contexts, yet the mechanisms supporting such cross-modal alignment remain poorly understood. In this work, we introduce a computational framework designed to model core aspects of human referential interpretation by integrating linguistic utterances with perceptual representations derived from large-scale, crowd-sourced imagery. The system approximates human perceptual categorization by combining scale-invariant feature transform (SIFT) alignment with the Universal Quality Index (UQI) to quantify similarity in a cognitively plausible feature space, while a set of linguistic preprocessing and query-transformation operations captures pragmatic variability in referring expressions. We evaluate the model on the Stanford Repeated Reference Game corpus (15,000 utterances paired with tangram stimuli), a paradigm explicitly developed to probe human-level perceptual ambiguity and coordination. Our framework achieves robust referential grounding. It requires 65\% fewer utterances than human interlocutors to reach stable mappings and can correctly identify target objects from single referring expressions 41.66\% of the time (versus 20\% for humans). These results suggest that relatively simple perceptual–linguistic alignment mechanisms can yield human-competitive behavior on a classic cognitive benchmark, and offers insights into models of grounded communication, perceptual inference, and cross-modal concept formation. Code is available at~\url{https://anonymous.4open.science/r/metasequoia-9D13/README.md}.}




\maketitle

\section{Introduction}

Effective cooperation between co-performers in joint activities depends on their ability to establish, maintain, and update a shared representation of common ground, which encompassed knowledge of the task, environment, and each other's capabilities. Achieving such common ground is cognitively demanding, even for humans, particularly under conditions of partial observability that can produce misinterpretations or coordination errors. Human interlocutors often begin with different conceptualizations of the same object, yet over repeated interactions they tend to converge on shared terminology through a process known as \emph{lexical entrainment} \cite{brennan1996conceptual}.

Lexical entrainment facilitates the formation of \emph{conceptual pacts}—temporary, partner-specific agreements about how to refer to objects or states in a shared context \cite{brown1958words}. Interlocutors establish these pacts by selecting referring expressions that are informative enough to disambiguate a target from alternative objects while avoiding unnecessary detail, consistent with Grice's Cooperative Principle \cite{grice1975logic}. This principle posits that humans typically adhere to the maxims of Quality (truthfulness), Quantity (informativeness), Manner (clarity and brevity), and Relation (task relevance). Conceptual pacts are flexible, maintained only with respect to a particular partner; introducing novel expressions outside this established common ground can induce \emph{partner-specific interference}, manifesting as slower responses, confusion, or reduced accuracy \cite{Trainin2025-tm}.

\begin{figure*}
\centering
\includegraphics[width=.9\textwidth]{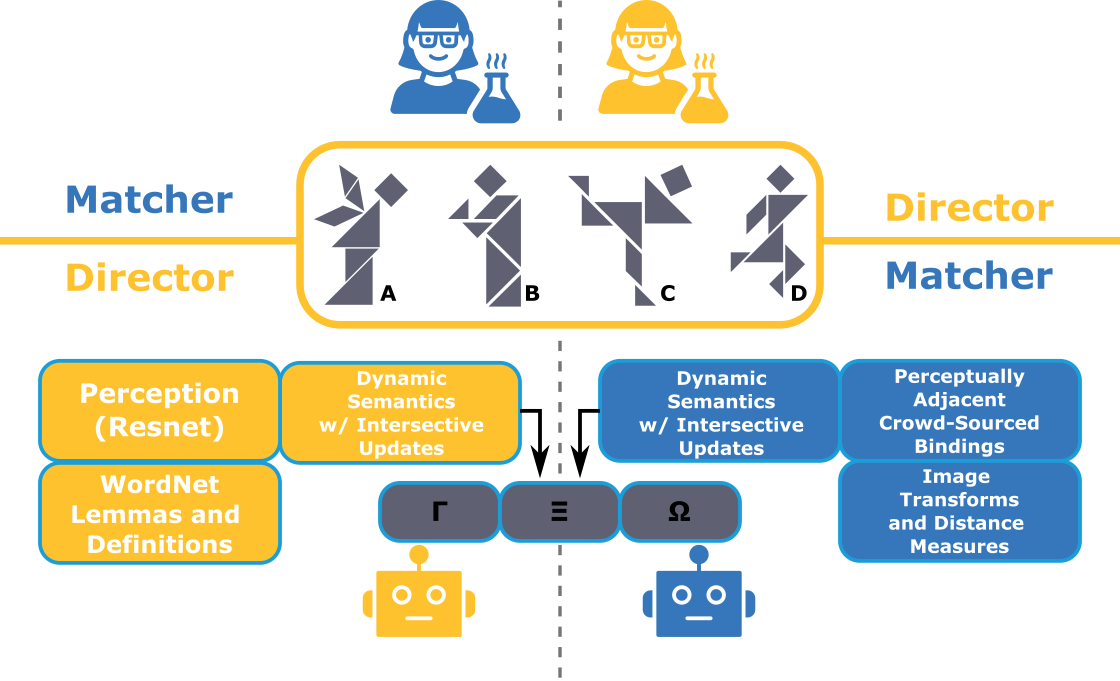}
\caption{An overview of the repeated reference game and our framework for lexical entrainment and common ground establishment. Our paper addresses the right-hand side of this figure, where the human serves as director and the AI as matcher. The sets $\Gamma, \Xi, \Omega$ maintain the current state of common ground, with $\Gamma$ containing finalized conceptual pacts, $\Xi$ the set of conceptual pacts under negotiation, and $\Omega$ containing any pacts which were rejected.}
\label{fig:framework}
\end{figure*}

In this paper, we present a computational framework for automatic lexical entrainment in a machine co-performer (MCP) engaged in the classic repeated reference game. In our experiments, the MCP serves as the matcher, tasked with aligning human-generated referring expressions $\varphi$ to the intended referent $r_{\varphi}$ bound to an object $o$ in the environment. Figure \ref{fig:framework} depicts the overall repeated reference game setup and our approach: the left-hand side shows a machine director with a human matcher (future work), while the right-hand side illustrates the current study of a human director and machine matcher. Both director and matcher have access to identical sets of $N$ randomly ordered abstract objects, $O = {o_0, o_1, \ldots, o_{n-1} }$, consisting of tangram stimuli that are deliberately challenging to describe. The director produces referring expressions $\Phi = {\varphi_0, \varphi_1, \ldots}$, which are then used to generate a set of referents $R = {r_{\varphi_i}, r_{\varphi_{i+1}}, \ldots, r_{\varphi_{i+(n-1)}} }$, each linked to a unique object via conceptual pacts, $r_{\varphi_k} \leftarrow o_j$, enabling the matcher to identify the target object $o_j$. Participants may use any speech acts necessary, but cannot share perceptual information outside of natural language.

Our MCP matcher is evaluated using the Stanford Repeated Reference Game corpus, containing over 15,000 director–matcher utterances~\cite{https://doi.org/10.1111/cogs.12845}
. While the MCP has access to the same tangram stimuli as human participants, these objects remain challenging, producing a cognitively demanding alignment task. The MCP leverages scale-invariant feature transforms (SIFT) \cite{lowe2004sift} to map crowd-sourced images to experimental stimuli and employs the Universal Quality Index (UQI) to quantify image similarity, thereby modeling perceptual alignment in a manner analogous to human visual comparison.

The main contributions of this work are:

\begin{itemize}
\item A novel formulation of common ground and conceptual pacts grounded in Update Semantics \cite{goldstein2019generalized}, capturing the dynamic and partner-specific nature of lexical entrainment.
\item A procedure for successful machine lexical entrainment based on this common ground representation.
\item Methods for improving alignment between human and machine perceptual spaces using sheaves constructed over SIFT features from crowd-sourced images, enabling the MCP matcher to map latent perceptual representations to symbolic referents.
\item Empirical evaluation on the Stanford open corpus of 15,000 director–matcher utterances, showing that the MCP achieves lexical entrainment with 65\% fewer utterances than humans, and correctly aligns a single referring expression 41.66\% of the time.
\end{itemize}

\section{Motivation, Background, and Related Work}
\label{motivation}

As AI and automated reasoning capabilities have advanced in such fields from genetics~\cite{bingham2022guide} and neurology~\cite{Bingham2025}, to controller state estimation for agriculture~\cite{bonsai} and embedded power infrastructure~\cite{9927846} there is an increasing need to develop the capacity for machines to perform less like automated tools and more like dynamic teammates, capable of taking on complex tasks with interdependence with other machine and human co-performers.  This new class of emerging Symbiotic AI \cite{wang2019symbiotic} is fundamentally different from prior AI applications, which primarily focused on monolithic AI systems capable of autonomy, and typically operating as a solitary machine in a non-social environment.  Next-generation AI systems must have the capability to reason socially and engage with team members in a fundamentally interdependent fashion, which requires the establishment, maintenance, and update/repair of common ground.

This capability is especially desirable with the rise of neurosymbolic AI, which combines the strengths of deep learning, including the ability to learn from experiences, with the ability to reason abstractly.  To fully leverage these advances neurosymbolic systems must be capable of linking latent spaces not only to symbolic logic, but also to natural language concepts present in the mind of human co-performers \cite{hamilton2022neuro}.

\subsection{Common Ground}

Common ground is a cornerstone to joint activities amongst humans.  Common ground, in this sense, is from the philosophy of language and refers to propositions, definitions, and other assumptions agreed upon by two individuals involved in discussion. Common ground during co-performance can be taken to be the set of jargon shared by co-performers that impacts task outcomes.  It is, in essence, a formalized shared model of communication. A core responsibility of a machine co-performer should be the active establishment and management of this common ground, and a representation of the context set that corresponds to this common ground.  We define a common ground $C$ as the context set of possible worlds $\{w_i, w_{i+1}, w_{i+2}, \ldots \}$ in which the conceptual pacts $\Xi = \{ r_a \leftarrow o_i, r_b \leftarrow o_j, \ldots\}$ indicating referent-object bindings are true.  

\begin{figure}
    \centering
    \includegraphics[width=0.7\textwidth]{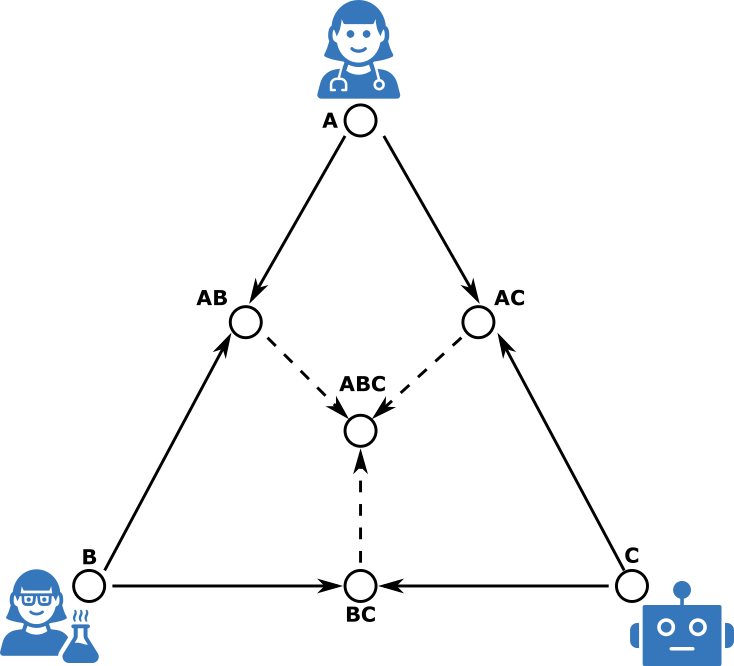}
    \caption{The symmetric simplicial sets of common ground that exist between human co-performers $A$ and $B$, and machine co-performer $C$.  Each pair of co-performers has their own common ground, representing shared understanding about themselves, their joint activity, and the environment, represented by the nodes $AB, AC, BC$ at the learned Wasserstein barycenters of mutual alignment.  Common ground $ABC$ is shared with all.}
    \label{fig:simplicial}
\end{figure}

The common ground between co-performers can be represented using category theory as symmetric simplicial sets in a barycentric coordinate system \cite{vince2025barycentric} as illustrated in figure \ref{fig:simplicial}. For $(n+1)$-performers we define the structure of their existing symbolic knowledge as a standard combinatorial n-simplex defined as the simplicial complex $S(n)$ where $n = {0, \ldots, n}$.  Each vertex in this simplex represents the labeled symbolic knowledge of one performer, shown as $A, B,$ and $C$. In between each of these representations at the Wasserstein barycenters of the simplex are the common grounds associated with any subset $m \leq n$ of the $n+1$ performers representing a sort of "average" of semantic understandings with ways to express homotopy and homology between concepts, without understanding the underlying topology of the languages in which common ground is established.  The process of establishing, maintaining, and repairing common ground is the process of estimating any barycenter of the simplex between co-performers.

Formal methods for the establishment of common ground provide new capabilities for MCPs.  Establishing common ground means that beliefs, assumptions, and intentions are shared among team members. While common ground needs to be constantly monitored, updated, maintained, and repaired, doing so provides a number of key properties, such as mutual predictability, ensuring that human and machine team members can maintain a shared picture of what’s happening in the world, and accurately predict co-performance outcomes.  Lexical entrainment is a key part of this process, allowing MCPs to generate symbolic mappings from latent spaces to symbols in knowledge bases, assumptions, beliefs, etc.  Lexical entrainment also allows the common ground to be formally inspected for inconsistencies, violated assumptions, or other errors, forestalling potential breakdown of team function \cite{klien2004ten}.

Automated lexical entrainment will allow machines to form conceptual pacts with human co-performers, reason about joint activity, to notify team members of impending failures \cite{Friedenberg2023-ml}, to understand and accept joint goals, to align their objective functions \cite{aguirre2020ai}, and signal when they are unable or unwilling to participate.  Lexical entrainment also establishes better predictability allowing for clarifying statements, assertions, and the ability of humans to assert directability. Specifying, it allow for dynamic objectives and commands in ways that can be understood by the machine and entered in an error free manner by humans.

Conceptual pacts resulting from lexical entrainment also allow humans to supplement machine limitations in perception and cognition, providing a model for joint activity and interdependence and supporting co-active design of the joint activity.  By representing joint activity in these formal manners, conceptual pacts allow MCPs to understand the goals of the human users and to measure and improve their own AI loyalty \cite{aguirre2020ai}, by using these conceptual pacts to engage in after action review \cite{dodge2021after} with models of human and machine intent \cite{billings2018aviation}.

\subsection{The Repeated Reference Problem}

The repeated reference problem~\cite{https://doi.org/10.1111/cogs.12845} is an exercise in common ground establishment found frequently in cognitive science and sociological literature.  We focus on the variant of the game associated with the Stanford open corpus of more than 15,000 utterances, which we use to evaluate our framework experimentally.

\begin{figure}
    \centering
    \includegraphics[width=.7\textwidth]{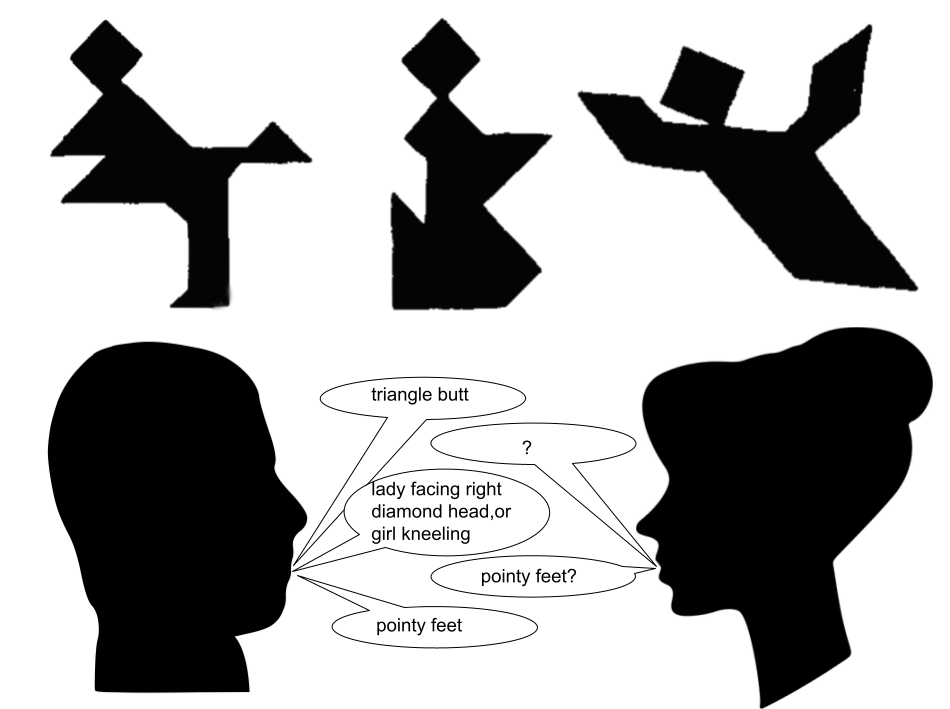}
    \caption{An example of the repeated reference problem, the director on the right, the matcher on the left.  The director issues an utterance, $\varphi$, indicating what they perceived the selected tangram stimuli to depict.  The matcher can either guess which tangram they believe the director is referring to, pose a clarifying question, or wait for the director to provide more information.  The illustrated example uses text from the open corpus.}
    \label{fig:game}
\end{figure}

The repeated reference problem involves a game played by two parties, a director and a matcher.  Both parties are provided with identical sets of tangram stimuli, but which have no labels and are in randomized, dissimilar, orders.  In this variant of the problem the director selects a tangram and produces an utterance, $\varphi$, in the form of a natural language sentence.  Their goal is to create concise utterances for the matcher that can be uniquely bound to individual objects in the set of tangram stimuli.  The matcher on the other hand must take the utterance and produce hypotheses as to which object is intended to be bound to the referent associated with $\varphi$.  The matcher then can propose a guess, issue a clarifying utterance, or wait for more guidance from the director in the form of another utterance, as seen in~\ref{fig:game}.

This is a canonically difficult game for human agents in the field of sociology and communication~\cite{https://doi.org/10.1111/cogs.12845}, and to date has not been addressed successfully by machines.  The matcher role, the subject of this paper, is particularly difficult due to the need to align with human perceptual spaces to understand the referent-object bindings implied by a referring expression $\varphi$.  The tangram stimuli used by this exercise is favored primarily because it is difficult for even human performers to align on perceptually.  We implement an MCP for the matcher role, aligning human and machine perception by using transformations of $\varphi$ to produce search terms for web-scraping, resulting in sets of crowd-sourced images related to the provided utterance. We detail this process further in Section~\ref{methods}.

\subsection{Dynamic Semantics}
\label{Sec:Dynamic}

We model the process of lexical entrainment over humans with a formalization based in dynamic semantics \cite{stalnaker2002common}.  Dynamic semantics is a perspective on natural language semantics emphasizing the growth of information over time.  It asserts that utterances during lexical entrainment are instructions to update an existing context with new information, resulting in an updated context.  Common ground, formally, consists of a set of propositions that all interlocutors take to be true, while also taking that all other interlocutors take them to be true.  These propositions are the aforementioned conceptual pacts, $\Xi$.

Under dynamic semantics, assertions and presuppositions interact in a dynamic fashion.  Propositions are presupposed if they are part of the pre-existing common ground.  New propositions can also be asserted, as a form of instruction to the interlocutor to update shared context with new information.  In the case of dynamic semantics, these can be formalized into update semantics using classical propositional logic \cite{goldstein2019generalized}.  We borrow Goldstein's notation, using $\varphi$ to indicate an utterance.  Each $\varphi$ is mapped to a an interpretation function $[\varphi]$ with a context change potential that takes a context as input, and produces a modified context as output.  A context $C$ is consistent with $\varphi$ iff $C[\varphi] = C$.

Similar to Stalnaker's original work on assertions \cite{stalnaker1978assertion}, we achieve dynamic semantic capabilities with update semantics. When an utterance $\varphi$ is issued by a director, we map this utterance to a set of worlds $\{w_i, w_j, \ldots \}$ representing assumptions about the shared environment with the function $\llbracket \varphi \rrbracket$.  The matcher takes their current understanding of the context $C$ modeled as a set of worlds representing the current common ground, and intersects it with $\llbracket \varphi \rrbracket$ to implement updating $C$ with $[\varphi]$.

\subsection{Possible Worlds Semantics for Epistemic Modals}
\label{Sec:Epistemic}
Deriving the context change potential function requires alignment of perceptual spaces.  Ideally, the MCP would be able to uniquely map an utterance $\varphi$ to the intended context change potential function using intersecting updates: $C \cap \{(r_{\varphi} \leftarrow o)\}$ indicating the the director intends the referring expression to communicate that object $o$ should be called $r_{\varphi}$.  As alignment of perceptual spaces is an unsolved problem, we instead allow the MCP to estimate the function $[\varphi]$ for any utterance $\varphi$.  To do so, it builds a hypothesized set of potential bindings implied by the referring expression $\varphi$ as the set $B = \{(r_{\varphi} \leftarrow o_i), (r_{\varphi} \leftarrow o_j), \ldots\}$. This indicates the context change potential function $[\varphi]$ is believed to be one of the bindings in $B$. We utilize traditional possible worlds semantics for epistemic modals might ($\might$) and must($\must$) to quantify over possible worlds. 
$\might\sigma$ is true at world $w$ iff $\sigma$ 
 is true in some world $v$ accessible from $w$. $\must\sigma$ is true at $w$ iff $\sigma$ is true for all worlds $v$ accessible from $w$

If the estimated meaning of $\varphi$ is one of the updates in $B$, then we say 
\begin{eqnarray*}
C[\varphi] &=& C \cap \might B \\
&=& C \cap \might \left\{(r_{\varphi} \leftarrow o_i), (r_{\varphi} \leftarrow o_j), \ldots\right\}\\
&=& C \cap \left\{ \might (r_{\varphi} \leftarrow o_i) \vee \might (r_{\varphi} \leftarrow o_j) \vee \ldots \right\}
\end{eqnarray*}

If $|B| = 1$ then additionally we can say

\begin{eqnarray*}
C[\varphi] &=& C \cap \must B\\
&=& C \cap \must (r_{\varphi} \leftarrow o_j)
\end{eqnarray*}

indicating potential lexical entrainment of part of the problem space. If $|B| = 0$ then the MCP matcher has failed to use the utterance to form hypothesized bindings and must wait for another from the director.  If $|B| > 1$ then the MCP matcher has a set of hypothesized bindings to choose from and needs additional information. Due to the limitations of working with a prerecorded public corpus our MCP was unable to ask clarifying questions of its own design.  In cases where $|B| > 1$, our MCP implementation waits for additional clarifying utterances in the data set to generate further object bindings, resulting in a strictly harder version of the problem.

To implement our MCP, we model the context of estimated common ground as the sets $\Gamma, \Xi,$ and $\Omega$, discussing update semantics further in section \ref{Sec:Calc}.  The set $\Gamma$ contains all established conceptual pacts which \textbf{must} be true, the set $\Xi$ contains the set of all conceptual pacts which \textbf{might} be true, and the set $\Omega$ contains the set of all conceptual pacts which have been rejected or disproven, including pacts of the form $\must \neg (r_{\varphi} \leftarrow o_i), \forall o_i \in O$ if the MCP fails to perceptually align on a given referring expression $\varphi$.  These negative pacts are, in essence, an agreement not to use the referent $r_{\varphi}$ to refer to any object.

\subsection{Perceptual Alignment}

Our procedure for the MCP matcher relies on building estimates of the context change potential function $[\varphi]$ by estimating the bindings implied by an utterance $\varphi$ using crowd-sourced images.  Because perception of tangram stimuli is varied, 
we model human perception using crowd sourced data which maps images to semantic labels. Using a set of images outside of the current tangram stimuli that are associated with semantic tags relevant to our corpus we queried popular search engine APIs with transformations of the utterance $\varphi$ as a means of estimating the meaning of the referring expression. Current works in this field focus on providing faithful results based on query parameters~\cite{9051800} and using these results to augment a user's understanding of real-world information~\cite{VISHWAKARMA2019217}. We utilize the Bing web image scraping API~\cite{bing_api} to submit our queries drawn from the Stanford corpus.

Bing's image suggestion utilizes trending image selection heavily to inform its suggestion model. This means that if a user who previously searched for a term similar to $\varphi$ selects an image, that image and other images that share common features become more likely to be recommended to future users.  We utilize Bing to estimate the context change potential implied by the director by using image matching to compare the set of crowd sourced images returned by transformations on $\varphi$ and use image matching techniques to compare them to our tangram stimuli.

Given a set of crowd sourced images resulting from $\varphi$, called $I_{\varphi}$, these image matching techniques provide us with distance metrics that allow us to reason that $\might(r_{\varphi} \leftarrow o_j)$ if the images in $I_{\varphi}$ are \emph{close} to $o_j$.

The MCP matcher uses metrics of image similarity to determine closeness of the crowd sourced images to the tangram stimuli.  Given some threshold $\epsilon$ and a similarity function $g(o_i, I_{\varphi})$, if $g(o_i, I_{\varphi}) > \epsilon$ then $\varphi \Rightarrow \might (r_{\varphi} \leftarrow o_i)$.  This provides a way to quantify how related the crowd sourced images are to our tangram stimuli images. The similarity function,  $g(o_i, I_{\varphi})$, can be implemented in a number of different ways, such as mean squared error, peak signal-to-noise ratio, structural similarity index, universal quality image index, spectral angle mapper, etc~\cite{923389}. In this work, we have found that universal quality image index (UQI)~\cite{995823} empirically provides the best results, out-performing all other methods tested by approximately 16\%.  We attribute the performance of UQI due to the fact this method predicts the probability of shared features as the primary similarity metric. This is ideal for determining if two pictures have different shapes but common features that a human may use to identify and label the content.

\section{Methods}
\label{methods}
In this section we discuss the core formalisms and procedures we use to implement an MCP matcher capable of solving the repeated reference game with a human co-performer, establishing common ground through artificial lexical entrainment.

\subsection{Query Construction for Web-Scraping}
As previously mentioned, since we do not have direct access to the intent of the director for $[\varphi]$, our MCP matcher utilizes the Bing web-search API to convert the extracted text to a set of images.  Submitting the raw utterance $\varphi$ as the search term yielded poor results, on par with random guessing, with accuracies only a little above 8\% in our evaluations.  Instead, we explored a number of transformations on $\varphi$ to improve the results of crowd-sourcing images with higher similarity.  By adding cues to the queries, such as appending the text "tangram figure", and by removing stop words, query initial accuracy improved by over 4x.

\begin{figure}
    \centering
    \includegraphics[width=.8\textwidth]{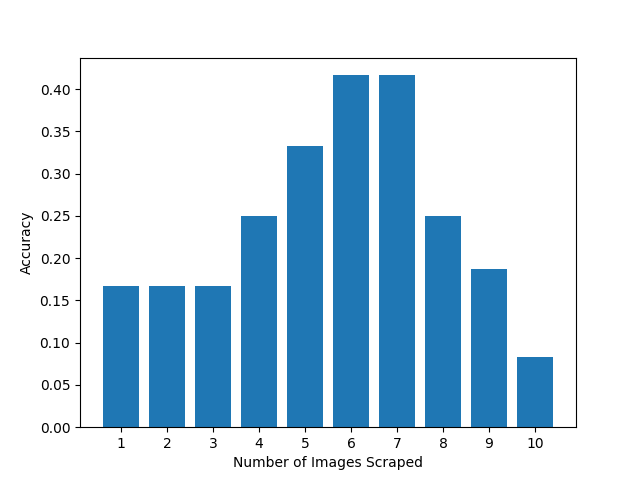}
    \caption{Changes in the accuracy of our bindings resulting from varying the number of images scraped by the MCP matcher.  These accuracy numbers represent achieving lexical entrainment on the referent in a single utterance, as a comparative measure.}
    \label{fig:n_images}
\end{figure}

For each of the queries formed above, another one of the hyper-parameters that had a large impact on the accuracy was the number of images scraped per query.  Estimating the intended context potential change function requires balancing two separate goals: firstly, to collect enough images to find representatives related to natural language utterances, but secondly to avoid scraping non-representative images.
As can be seen in Figure~\ref{fig:n_images}, the number of images utilized has a strong impact on the matcher. After 7 images, the images provided by the Bing API had a strong destabilizing effect on our decision spaces. This was due to the fact that after 7 or so results, many queries return a generic image of a solved, square tangram, which almost always matches with the same tangram in our object set.

\subsection{Image Matching}

After scraping candidate images $I_{\varphi}$ from the web, the MCP matcher then prepares a pipeline computing the similarity of these images to the tangram stimuli objects, shown in figure \ref{fig:processing}.  This pipeline aligns and normalizes the images, and attempts to extract and compare relevant shapes and features from the images to assess the comparative distance of each of the tangram objects to $I_{\varphi}$.

We intentionally employ classical perceptual similarity measures rather than end-to-end learned representations in order to preserve interpretability and to better align the model’s internal operations with established accounts of human perceptual comparison.

\subsubsection{Image Alignment}
The MCP first attempts to align relevant features in the scraped images, utilizing utilizing SIFT homographies~\cite{https://doi.org/10.48550/arxiv.2104.11693}. Similar to the sliding windows used in convolutional neural networks, this method starts by segmenting the images into 3x3 kernels. These kernels are compared and key points are matched based on a Gaussian difference method~\cite{Lindeberg:2012}. This algorithm is scale as well as rotational invariant. This makes it ideal for our application, where images may represent a similar subject from different perspectives, helping to align human and machine perception. 

\subsubsection{Image Comparison}

\begin{figure}
    \centering
    \includegraphics[width=.9\textwidth]{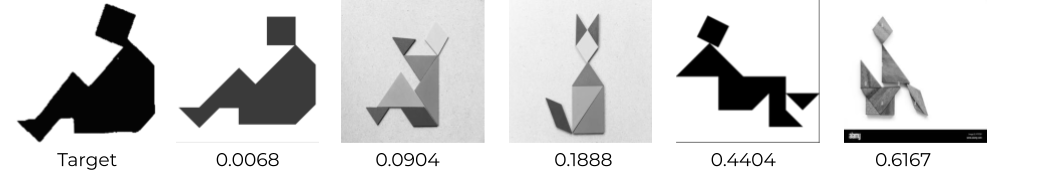}
    \caption{An example of the distances of the closest 5 scraped photos to the target. The query text for this is "tangram figure sitting and looking". \textit{It should be noted that all queries were manually checked to ensure that the scraped images were unique from the tangram figure.}}
    \label{fig:top_k}
\end{figure}

After the images are aligned, the images are replicated and subjected to a number of rotational transforms, as well as being replicated with grey-scale values inverted. Doing this helps to generalize our results, and raised the accuracy by approximately 8\%. These copies, along with the original picture, are then assessed based on how close they are to the given tangram stimuli by applying UQI~\cite{995823}. UQI is a normalized image quality measure which utilizes approximating the noise need to transform one image into another. This measure is applied to each cross product between the scraped images and the tangrams. An example of these outputs can be seen in~\ref{fig:top_k} for a single tangram compared to its top five results. 

\subsection{Formalizing Common Ground Establishment}
\label{Sec:Calc}

To model the process of common ground establishment through lexical entrainment, we define the context representing the common ground as the category of sets whose objects are the worlds consistent with three sets we use in the MCP's to model the common ground: $\Gamma, \Xi,$ and $\Omega$. $\Gamma$ is the set of conceptual pacts (represented by bindings) that the MCP believes must be true.  $\Xi$ represents the sets of bindings the MCP hypothesizes might be true based on utterances and perceptual alignment.  Finally, $\Omega$ represents the sets of bindings the MCP has determined must be false, including those bindings to referents the MCP was unable to resolve satisfactorily with image matching. The common ground is thus context $C$ of worlds $\{w_0, w_1, \ldots \}$ consistent with $\Gamma \cap \Xi \cap \Omega$.

Formally, when we estimate $C[\varphi]$ as $C \cap \might B$ we update our model such that \[ \Xi = \Xi \cap \might B\].  When any referring expression results in an unambiguous context change potential of the form $C[\varphi] = \must (r_{\varphi} \leftarrow o_i)$ we update our model such that:

\begin{eqnarray}
    \Gamma &=& \Gamma \cap \must (r_{\varphi} \leftarrow o_i)\\
    \Xi &=& \Xi \setminus \might \{ (r_{\varphi} \leftarrow o_j | o_j \in O \}\\
    \Omega &=& \Omega \cap \must \neg \{ (r_{\varphi} \leftarrow o_k | o_k \in O \mathrm{\ where \ } o_k \neq o_i\}    
\end{eqnarray}

Lastly, when perceptual alignment on $\varphi$ results in $B = \emptyset$ we update our model such that:

\begin{eqnarray*}
    \Omega &=& \Omega \cap \must \neg \{ (r_{\varphi} \leftarrow o_k | o_k \in O\}    
\end{eqnarray*}

We propose that for the repeated reference game, successful lexical entrainment on a referent $r_{\varphi}$ and an object $o$ indicates that the set of agreed upon conceptual pacts, $\Gamma$, contains the referent-object binding $(r_{\varphi} \leftarrow o)$.  Common ground is established, and lexical entrainment has succeeded when $\Gamma$ contains a unique referring expression $(r_i \leftarrow o), \forall o \in O$ with $\Xi = \emptyset$.  The state of $\Omega$ does not impact establishment, and is merely used to "memorize" unhelpful referents to avoid their use in the future.

The process of establishing common ground between interlocutors uses the theory from dynamic semantics referenced in section \ref{Sec:Dynamic}.  Dynamic semantics provides a framework where the meaning of a sentence is a richer concept than under static semantics (where knowing a sentence provides its meaning) \cite{goldstein2017informative}.  Dynamic semantics approaches the meaning of a sentence as a rule for learning whether or not the sentence is true.  These rules take the information state of any interlocutor (also known as it's context, $C$) applying an update to the context resulting from $\varphi$ using the function $[\varphi]$ defined in section \ref{Sec:Epistemic}.

\section{Experimental Methods and Results}

We evaluated our MCP matcher on Stanford's public corpus for the repeated reference game \cite{https://doi.org/10.48550/arxiv.1912.07199} as the basis of our experiments. The data set represents pairs of human deciders and matchers exchanging unique recorded utterances, each with an associated intended target object. 
 The data sets provide the time taken to make an utterance, the utterance itself in text form, and the associated visual tangram stimuli that were shown to the subjects.  Removing the matcher utterances, we exposed our MCP matcher to the more than 8,000 decider utterances.  The MCP matcher attempted to estimate context potential change functions for each utterance through web-scraping, and to resolve the utterance successfully into a binding rule using UQI similarity measures with varying values for the decision parameter $\epsilon$.

\subsection{Transformations on $\varphi$}

As the data set contains natural language text produced by human subjects, many of the tokens present in the utterances do not contribute to the quality of the context potential change function estimation, and can even hinder the matcher's ability to lexically entrain. This includes the many common prefixes such as "one that looks like".  By removing common stop words that make up such prefixes, as well as any tokens not representing a potential noun, verb, or conjunction we were able to dramatically improve our estimated context potential change functions. The difference of a "really tall man" and "tall man" does not contribute towards perception alignment, and often degrades the quality of the MCP matcher.  We implemented these transformations on stop words, and non-contributing parts of speech using  the spacy library for Python~\cite{spacy2}.

The final preprocessing steps taken on utterances was to attempt to detect and normalize spelling deviations.  Subjects in the experiment used informal language and spellings, and occasionally submitted typos.  We again leveraged the Spacy library to automatically normalize these results, improving outcomes.

\begin{figure*}[h]
    \centering
    \includegraphics[width=.95\textwidth]{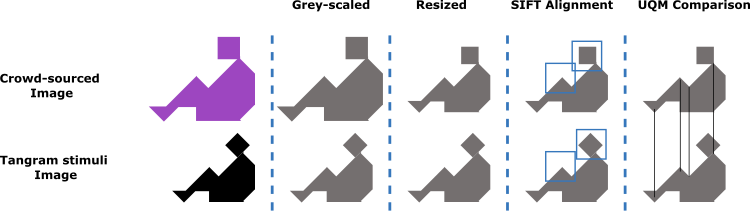}
    \caption{Processing of scraped images to conform with tangram data set from original to comparison measure. The use of grey-scale is in order to prevent color disagreements from disproportionately effecting results.}
    \label{fig:processing}
\end{figure*}

Like the text, the tangram images needed changes to assist with the matching algorithm. The images of the tangrams were flood filled, removing any artifacts that existed from the copying process. These black and white images were then resized to all consistent with each other, and the tangram stimuli. All samples were down sized to be 300x300, as shown in the final step of in Figure~\ref{fig:processing}, which summarizes our pipeline and its steps. 

\subsubsection{Image Distance Measure}
As explained before, we tested a multitude of different metrics for assessing the difference between two images. The measures we tested are mean squared error, mean absolute error, peak signal-to-noise ratio, structural similarity index, Erreur Relative Globale Adimensionnelle de Synthèse, Spatial Correlation Coefficient, Relative Average Spectral Error, Spectral Angle Mapper, Visual Information Fidelity, and Universal Quality Image Index~\cite{995823}.

In order to appraise which distance measure performed best in our application, we repeated the tests documented in Table \ref{tab:top_1_1_shot} with all measures previously listed. Further, as some of the aforementioned contain self-aligning algorithms as part of their measure, the test was also repeated with all measures with, and without pre-alignment. From these tests, we found that the Universal Quality Image Index (UQI) with SIFT alignment out performed all others by about 16\% accuracy.

\subsection{Implementation}
All experiments were done in Python 3 utilizing the the bing API~\cite{bing_api} to web-scrape crowd sourced images representing referring expressions. All times were taken using the time features in Python. These experiments were conducted on a MacBook Pro (15-inch, 2017) running 3.1 GHz Quad-Core Intel i7. No GPU were utilized for these experiments.

\subsection{Metrics}

In order to understand our success at lexical entrainment, we examine our resulting common ground outcomes from a set of diverse metrics to characterize the quality of our solution.  To understand the quality of our estimation of the context change potential function we examine the top-k accuracy achieved by our methods.  These results are summarized in table \ref{tab:top_1_1_shot}.  Unfortunately, due to lack of a record of the hypotheses a human holds as a result of an utterance, we are unable to compare our results to human matchers beyond top$_1$ accuracy, which we estimate on the basis of successfully lexically entrained referrants from a single utterance.  There were no instances of this in the public corpus.  Our implementation of the MCP matcher, however, was able to achieve lexical entrainment after one utterance $41.66\%$ of the time.

\begin{table}
    \centering
    \resizebox{.6\textwidth}{!}{
    \begin{tabular}{c|c|c}
        top$_k$ & Human & matcher  \\
        \hline
        \hline
         $k=1$ & 20.00 & 41.66\%\\
         \hline
         $k=3$ & N/A & 63.01\%\\
         \hline
         $k=5$ & N/A & 83.56\%\\
    \end{tabular}
    }
    \caption{Top-k accuracy of the game when only the first director's phrase is used. The human matcher got none of the tangrams correct with just one utterance from the director. This is starkly contrasted by our matcher, which was able to get 41.66\% correct with just one phrase from the director.}
    \label{tab:top_1_1_shot}
\end{table}

When lexical entrainment cannot be established with one utterance, we allow the matcher to propose additional hypotheses using a softmax function to convert the latent space of computed distances $g(o_i, I_{\varphi})$ into a probability distribution of hypotheses, with our decision criteria $\epsilon$ set to provide the indicated number of top hypotheses.  Allowing the MCP to utilize three hypotheses improved the rate of lexical entrainment from a single utterance from 41.66\% to 63.01\%, and utilizing five hypotheses resulted in lexical entrainment $83.56\%$ of the time.

\subsubsection{Speed of lexical entrainment} In addition to our success rate at entrainment with a single utterance by varying our closeness decision criteria $\epsilon$, we also measured the speed at which lexical entrainment was achieved, both in terms of cognition/reasoning time (shown in column 2 of table \ref{tab:time_and_uterances}), and the total number of utterances needed to reach our stopping criteria (shown in column 3 of table \ref{tab:time_and_uterances}).  As can be seen from our results, the MCP matcher out performed the user in terms of wall clock time, but more importantly, requiring on average only 65\% of the number of exchanged utterances, reaching entrainment on all tangrams with an average of $1.78$ utterances per object, vs. $2.73$ for human performers.  While actual speed of cognition may seem to be an unfair comparison, it is important to note in this work as it represents the difficulty humans experience at creating utterances, and more importantly, at interpreting utterances.  Reports on trials with the Stanford corpus, and other similar data sets place a heavy emphasis on the time taken for a given exchange, and further to establish lexical entrainment for a given tangram.  Achieving fast lexical entrainment with machine assistance is especially desirable in critical co-performance, where the lack of established common ground can prove detrimental to safety-critical, and even life-critical joint activities, such as triage, search and rescue operations, and other crisis decision making situations currently seeking to employ symbiotic AI to improve outcomes.

The matcher’s ability to converge with fewer utterances should not be interpreted as superior communicative competence, but rather as a consequence of relying on external perceptual regularities that human interlocutors typically negotiate interactively. This highlights a tradeoff between interactive grounding and perceptual bootstrapping available to a machine co-performer.

\begin{table}
    \centering
    \resizebox{.75\textwidth}{!}{
    \begin{tabular}{c|c|c|c|c}
                    & \multicolumn{2}{c}{Time (ms)} & \multicolumn{2}{c}{Utterances Needed} \\
         Tangram ID & Human & matcher &            Human & matcher~\footnote{This is the average when the matcher gets the correct answer.} \\
         \hline
         \hline
         A & 31737 & 1.2 & 2.5 & 1  \\
         \hline
         B & 21156 &  7.8 & 3.75 & 1 \\
         \hline
         C & 15311 & 3.3 & 2.5 & 2.3 \\
         \hline
         D & 27794 & 0.4 & 2.4 & 1 \\
         \hline
         E & 16614 & 2.9 & 2.4 & 1 \\
         \hline
         F & 50496 & 14.1 & 2.5 & 2.3 \\
         \hline
         G & 21756 & 2.1 & 2.4 & 2.5 \\
         \hline
         H & 26559 & 1.8 & 2.4 & 1 \\
         \hline
         I & 37634 & 2.4 & 2.4 & 1 \\
         \hline
         J & 37392 & 2.2 & 2.4 & 2.3 \\
         \hline
         K & 60380 & 2.9 & 4.8 & 1 \\
         \hline
         L & 42110 & 5.1 & 2.3 & 5 \\
         \hline
         Average & 32411.58  & 3.9 & 2.73 & 1.78\\
    \end{tabular}
    }
    \caption{A comparison of the performance between the human matcher from the data set  to our matcher give the same input phrases (utterances) from the director.}
    \label{tab:time_and_uterances}
\end{table}

\section{Conclusions}
The lack of automated solutions to the problem of lexical entrainment in the literature makes it difficult to compare our methods to prior work, as indeed this solution is the first in the literature that we, or our colleagues in cognitive and social science are aware of.  As such we believe these results are novel, and we present them in the context to the current state of the art achievable by human co-performers. Our current results are also limited in what can be inferred from the available public corpora.  Current corpora do not record the space of hypothesized lexical entrainments implied by an utterance that humans possess in their minds, and the intent of director actions are likewise not recorded, yielding no dependable ground truth by which to explicitly compare our estimation of the context change potential function. For example, there are instances where the human matcher asked a question to the director, like "pointy feet?". It is impossible to assert that they asked this because they believed that the correct tangram did, or believed it did not have pointy feet, where as with our MCP, the set of all held beliefs is extant in $\Xi$, explicitly. Within these limitations, our solution represents a first of its kind capability, showing successful lexical entrainment in less time, with fewer utterances than human performers are achieving, when the MCP is in the matcher role. This provides strong evidence that our matcher is superior when compared to humans. 

This work demonstrated the use of dynamic semantics with web-scraped crowd sourced images to estimate the context change potential function to implement an MCP matching agent, which approximates the capacity of a human matcher to achieve lexical entrainment on the repeated reference problem. We outlined the hyper-parameters, techniques, and algorithms required by our matcher and demonstrated that this method achieves the first known example of an automated solution to the repeated reference problem, for the matcher position using publicly available data, achieving comparable or more sample-efficient performance under the same informational constraints as human capability for our evaluation.  We make our code available on github, under an open source license at ~\url{https://anonymous.4open.science/r/metasequoia-9D13/README.md}.

\section{Discussion and Future Work}
\label{discussion}

While we have tested our methods in the limited domain of the repeated reference game, we believe these results can be substantially generalized to apply to more realistic problems of common ground establishment and lexical entrainment.  Furthermore, we believe we can improve our initial results and accuracy metrics by utilizing a live exercise with active human subjects, rather than pre-recorded exchanges, affording our MCP matcher the ability to pose its own clarifying questions.  Many utterances provided by the director could not be successfully transformed into hypothesis generating queries (i.e. the resulting $B$ for the image set $I_{\varphi}$ has cardinality 0, given $\epsilon$. For example, the utterance "zig zag with square on top" does not produce faithful results when submitted to the search engine when looking for tangrams. Often, when the query was too far away from a concept the engine could grasp, it would simply return a solved square tangram. As mention before in Section~\ref{methods}, representative images of a solved square tangram hinders the capacity of the matching algorithm considerably. It also does not provide any meaningful information in terms of being a geometric interpretation of the director's utterance. This means that for certain phrases, this algorithm will not work. That being said, the human matcher also did not perform well on these, and requiring further clarification, which our MCP matcher was able to utilize.

\section*{Funding Statement}
Work was funded by Galois Inc.

\section*{Conflict of Interest}
Author reports conflicts with Galois Inc., John Deere, Iowa State University, Rutgers University, and Technion University. 

\section*{Data Availability}
The data utilized is publicly available at https://doi.org/10.1111/cogs.12845~\cite{https://doi.org/10.1111/cogs.12845}. 

\section*{Declaration of Usage of AI}
Author reports AI was used for checking grammar and spelling, and not in a generative fashion for writing or creating sections.


\bibliography{sn-bibliography}

\end{document}